\documentclass[sigconf,nonacm,screen]{acmart}
\renewcommand\footnotetextcopyrightpermission[1]{}

\makeatletter
\def\acmConference@shortname{\textbf{Memor-E}}%
\def\acmConference@date{}%
\def\acmConference@venue{}%
\def\acmConference@shortauthors{}%
\makeatother

\AtBeginDocument{%
  }

\usepackage{tcolorbox}
\usepackage{balance}
\usepackage{needspace}
\usepackage{graphicx}
\usepackage{tabularx}
\usepackage{xurl}
\usepackage{eso-pic}
\usepackage{wrapfig}
\usepackage{placeins}
\usepackage{float}

\begin{document}

\title{MEMOR-E: In-Context and Fine-Tuned LLM Personalization \\for Alzheimer's Assistive Robotics}

\author{Maissa Abir Smaili}
\affiliation{%
  \institution{Istanbul Medipol University}
  \city{Istanbul}
  \country{Türkiye}}
\email{maissa.smaili@std.medipol.edu.tr}

\author{Eren Sadikoglu}
\affiliation{%
  \institution{Arizona State University}
  \city{Tempe}
  \country{USA}}
\email{esadikog@asu.edu}

\author{Ransalu Senanayake}
\affiliation{%
  \institution{Arizona State University}
  \city{Tempe}
  \country{USA}}
\email{ransalu@asu.edu}

\begin{abstract}
Alzheimer's disease is a neurodegenerative disorder marked by progressive declines in memory and language that reduce independence in daily life, motivating socially assistive robotic support.

This paper presents \textbf{MEMOR-E}, a mobile quadruped robot with an interactive tablet interface that assists patients and caregivers through medication reminders, routine guidance, memory oriented interactions, and companionship.

We evaluated the feasibility of fine tuning large language models (LLMs) to emulate stage consistent cognitive behavior and interpret responses across standard neuropsychological language tasks, using audio transcriptions from 235 Alzheimer's patients and synthetically generated healthy controls.

We also report findings on using in context learning (ICL) in LLMs, where a second LLM produced domain and severity level cognitive error summaries. Our results show that {MEMOR-E} can generate stage aware, non diagnostic cognitive summaries that support personalized assistive interactions, while explainable AI mechanisms translate model outputs into transparent, human readable evidence to enable caregiver oversight and trustworthy human robot interaction.
\end{abstract}



\maketitle

\section{Introduction}

Alzheimer's disease affects more than 55 million individuals worldwide and it is characterized by progressive decline in memory, language, and executive functions~\cite{jack2018framework}. These impairments increase reliance on caregivers and family members for medication management, daily routines, and emotional support. These demands place substantial emotional and logistical burdens on families and healthcare system.

The recent advances in robotics and artificial intelligence have enabled the social assistant robots that's compliments human care given by offering reminders and companionship~\cite{bemelmans2012scoping}. The cognitive simulation prior work demonstrates that pet robots can reduce stress improve engagement, and increase quality of life for Alzheimer's patients. However, many existing systems are either stationary, purely affective, or lack integration with modern language based reasoning systems.

\begin{figure}[t]
  \centering
  \includegraphics[width=0.85\linewidth]{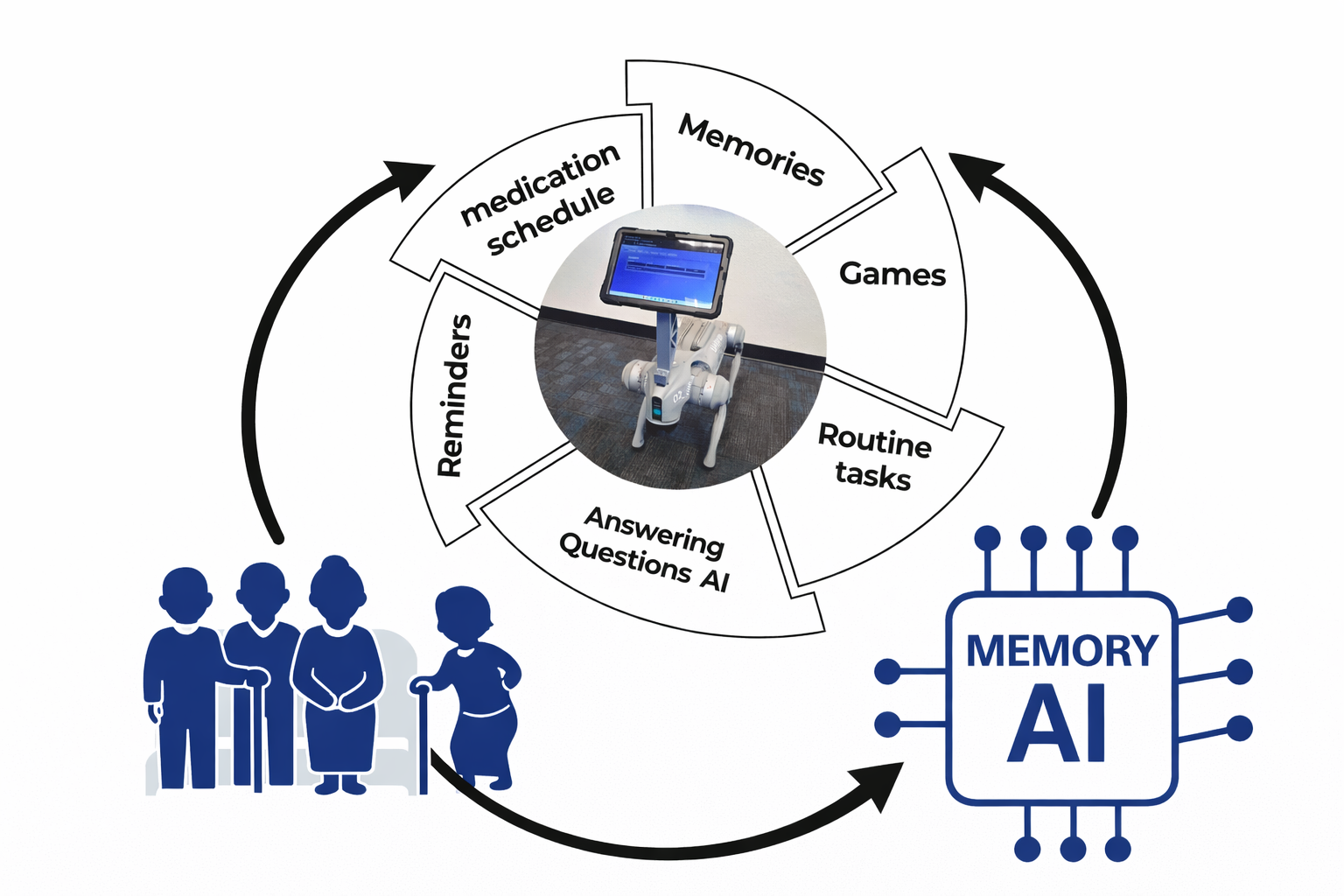}
  \caption{Alzheimer's helper robot dog MEMOR-E with interactive screen and functional modules.}
  \label{fig:system}
\end{figure}

\begin{figure}[t]
\centering
\includegraphics[width=0.5\linewidth]{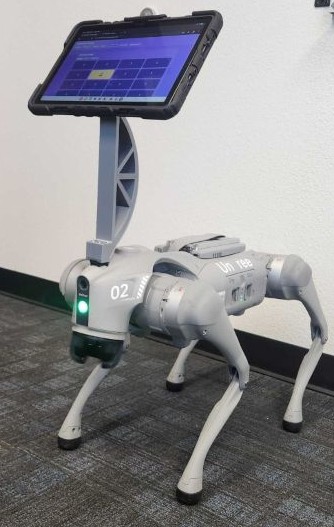}
\caption{MEMOR-E assistive robot with a head mounted tablet interface used for delivering reminders, cognitive games, and memory oriented interactions.}
\label{fig:memo2}
\end{figure}

We introduce MEMOR-E, a mobile quadruped robot with a head-mounted tablet providing context-aware assistance through visual prompts, AI-assisted dialogue, reminders, medication support, cognitive games, and memory cues using pictures and videos with descriptions. MEMOR-E physically approaches the user and delivers information at the point of need. The system also explores the feasibility of large language models to simulate stage consistent Alzheimer's cognitive behavior and generate interpretable summaries of memory performance for assistive interaction, not diagnosis. The contributions are as follows:

\textbf{System Design with Explainable AI Support:} An integrated hardware and software platform combining a quadruped robot, tablet interface, autonomous navigation, caregiver connectivity, and explainable AI to produce transparent, human readable interpretations of cognitive task outcomes.

\textbf{Alzheimer's Stage Adaptive Interaction:} A simplified graphical interface aligned with cognitive challenges across Alzheimer's disease stages, derived from clinical assessments and supported by interpretable feedback for caregiver oversight.

\textbf{LLM-Based Cognitive Analysis:} A feasibility study using large language models to emulate stage consistent patient responses, categorize cognitive errors across neuropsychological tasks, focusing on interpretable domain level summaries rather than predictions.

\section{Related Work}

\paragraph{Socially assistive robotics for dementia.}
Socially assistive robotics has been widely explored in dementia care~\cite{rossi2017users}, including humanoid robots' conversational agents, and Pet robots such as Paro have demonstrated positive effects on mood, stress reduction, and social engagement in elder care environments. Some of them are focusing on task reminders and cognitive games. However, the mobility and personalization are limited~\cite{chapman1995narrative}.

\paragraph{Language and memory assessment in Alzheimer's disease.}
Language based cognitive assessment has also gained attention, particularly using speech transcripts from picture description, for verbal fluency, and story recall tasks~\cite{perret1974fluency}. Large datasets such as talk bank dementia bank~\cite{becker1994natural} provide annotated speech samples that reveal linguistic markers of Alzheimer's disease, including reduced lexical diversity, increased disfluencies, and narrative fragmentation. These tasks prop the memory executive function and syntactic organization. The datasets enabled computational analysis of dementia related language decline.

\paragraph{Large language models in healthcare and HRI.}
LLMs have recently been explored for clinical text analysis, patient simulation, and explainable summarization. While promises and concerns remain regarding diagnostic misuse. In this work, LLMs are used strictly for behavioral simulation and interpretability, support, and assistive interaction rather than medical decision making.

\section{Methodology}

MEMOR-E follows a two-step, privacy preserving framework for cognitive severity detection and assistive feature planning. The system integrates transformer-based language modeling, explainable artificial intelligence (XAI), and a local large language model (LLM) within a mobile robotic platform.

\subsection{Hardware and Software Architecture}

MEMOR-E is deployed on a Unitree Go2 quadruped robot, selected for its stable locomotion, compact indoor mobility, and socially acceptable embodiment in assistive environments. A head-mounted touchscreen tablet serves as the primary interaction interface, delivering reminders, cognitive exercises, and visual prompts.

The system operates under ROS~2, using Nav2 for autonomous navigation and safe indoor mobility. An Intel RealSense RGB-D camera supports obstacle avoidance and basic user awareness. All computation is performed locally, enabling on-device execution of the Longformer classifiers and a Qwen~2.5 (7B) LLM without transmitting patient data externally.

The architecture follows a closed-loop structure:
\begin{itemize}
    \item Perception informs cognitive state estimation.
    \item Transformer-based models compute task-specific severity signals.
    \item XAI-derived summaries guide downstream reasoning.
    \item A local LLM maps severity signals to assistive feature recommendations.
    \item The robot executes navigation or tablet-based interaction accordingly.
\end{itemize}

\begin{figure*}[!tp]
\centering
\includegraphics[width=0.95\textwidth,trim=0.5cm 18cm 0cm 0cm,clip]{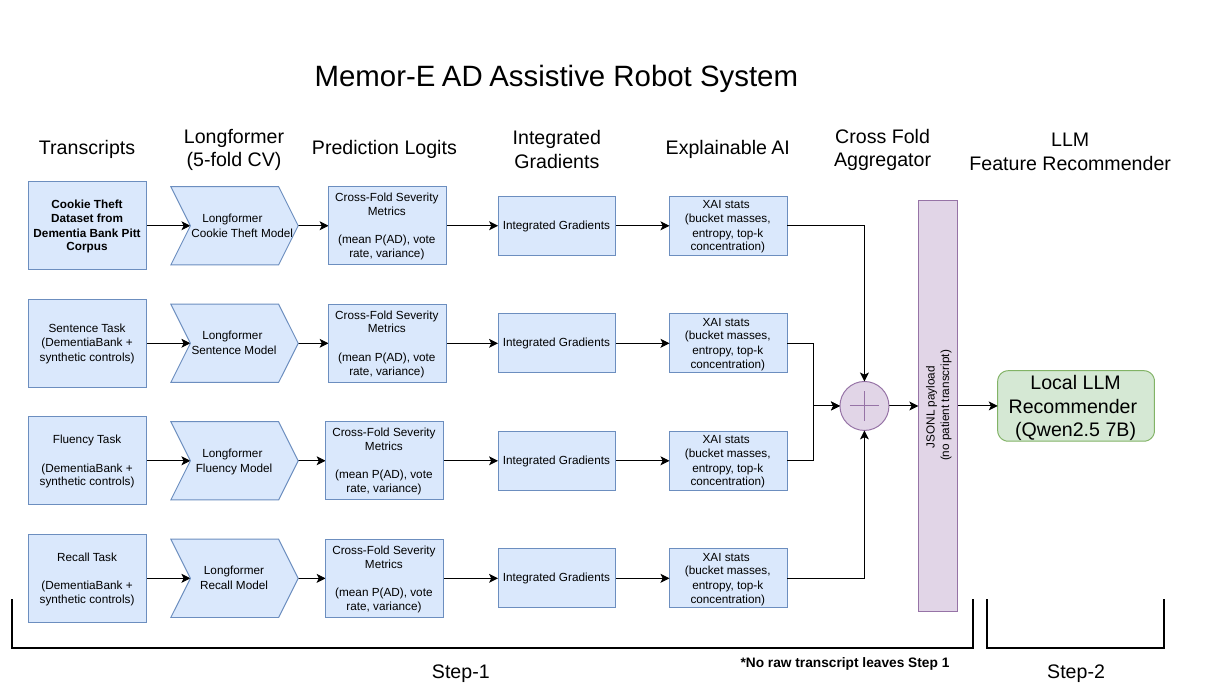}
\caption{MEMOR-E pipeline: transcript-based Longformer classifiers produce task-specific severity signals; Integrated Gradients based XAI aggregates privacy-preserving bucket statistics; a local Qwen 2.5 LLM maps statistics to non-diagnostic assistive feature plans; the robot executes tablet interactions and navigation.}
\label{fig:architecture}
\end{figure*}

Figure~\ref{fig:architecture} illustrates the complete MEMOR-E processing pipeline.
MEMOR-E prioritizes system-level transparency and controllability over full autonomy, explicitly constraining learned components within a closed-loop, human-supervised assistive architecture.

\subsection{Cognitive Tasks and Datasets}

We evaluated MEMOR-E across four standard neuropsychological language tasks derived from the TalkBank DementiaBank Pitt~\cite{becker1994natural}, which is a fully anonymized dataset. We followed their guidelines while working on the dataset.

\subsubsection{Cookie Theft Picture Description}

Participants describe a complex visual scene depicting children stealing cookies while their mother is distracted. This task evaluates semantic memory, visual attention, and narrative coherence. Both Alzheimer's disease (AD) patients and real healthy controls are available for this task.

\subsubsection{Story Recall}

Participants recall the ``George and Melanie'' story under three conditions:
Immediate recall, Delayed recall, and Comprehension questions.

This task evaluates episodic memory and memory consolidation. AD samples are real; healthy controls were synthetically generated to balance class distributions.

\subsubsection{Verbal Fluency}

Participants produce:
\begin{itemize}
    \item Animal names (semantic fluency),
    \item Words beginning with the letters F or S (phonemic fluency).
\end{itemize}
This task assesses lexical retrieval and executive control.

\subsubsection{Sentence Construction}

Participants generate:
\begin{itemize}
    \item A sentence containing a target word (e.g., \textit{tree}),
    \item A sentence containing three given words (e.g., \textit{doctor, chair, child}).
\end{itemize}
This task captures syntactic planning and semantic integration.

\subsubsection{Synthetic Healthy Controls}

For the Recall, Fluency, and Sentence tasks, healthy control samples were generated using a large language model to balance class distributions. These samples were used exclusively for classifier training and are not interpreted as evidence of real-world generalization~\cite{lehman2021bias}. Dataset composition is summarized in Table~\ref{tab:datasets}.

\begin{table}[h]
\centering
\small
\caption{Datasets used in the multi-task experimental setup. Tasks involving synthetic healthy controls are interpreted with caution due to potential distributional artifacts.}
\setlength{\tabcolsep}{4pt}
\begin{tabularx}{\columnwidth}{lccccX}
\hline
\textbf{Task} & \textbf{\#AD} & \textbf{\#HC (real)} & \textbf{\#HC (syn)} & \textbf{Total} & \textbf{Notes} \\
\hline
Cookie Theft & 309 & 243 & 0 & 552 & Real AD and real healthy controls. \\
Recall & 263 & 0 & 200 & 463 & Synthetic HC generated using an LLM. \\
Fluency & 235 & 0 & 235 & 470 & Synthetic HC generated using an LLM. \\
Sentence & 236 & 0 & 236 & 472 & Synthetic HC generated using an LLM. \\
\hline
\end{tabularx}
\label{tab:datasets}
\end{table}

\subsection{Step 1: Explainability-Driven Cognitive Signal Extraction}

\subsubsection{Longformer-Based Classification}

We employed \url{allenai/longformer-base-4096} to accommodate long clinical transcripts. For each task, an independent binary classifier was trained:
\[
F_{\text{task}}: \text{transcript} \rightarrow P(\text{AD})
\]

Models were evaluated using stratified five-fold cross-validation with binary cross-entropy loss. Metrics included Accuracy, F1-score, AUC, Sensitivity, and Specificity. For each subject, we computed:
\begin{itemize}
    \item Mean AD probability across folds,
    \item Cross-fold vote rate,
    \item Probability variance as a stability measure.
\end{itemize}

\subsubsection{Explainable AI Profiling}

Token-level attributions were extracted using Integrated Gradients~\cite{sundararajan2017axiomatic} (Captum). Integrated Gradients is an attribution method that estimates how strongly each input feature contributes to the model's prediction by integrating gradients along a continuous path from a baseline input to the actual input, thereby capturing the directional influence of each token on the output logit while satisfying axiomatic properties such as sensitivity and implementation invariance. Because Longformer uses byte-pair encoding (BPE), subword tokens were reconstructed into word-level units prior to attribution analysis.

Attributions were grouped into \emph{linguistically motivated buckets} to enable privacy-preserving aggregation while capturing clinically relevant speech patterns. Specifically, we bucketed tokens into:
\begin{itemize}
    \item Disfluency and annotation markers (e.g., fillers, repairs, CHAT tags),
    \item Lexical content tokens (open-class words),
    \item Punctuation,
    \item Short subword fragments,
    \item Special model tokens.
\end{itemize}

From these, we derived privacy-preserving statistics:
\begin{itemize}
    \item Disfluency-to-content ratio,
    \item Normalized bucket mass distribution,
    \item Evidence entropy,
    \item Attribution concentration measures.
\end{itemize}
Importantly, this explainability pipeline ensures that no raw patient transcripts are exposed beyond Step 1 processing, preserving privacy while enabling interpretable cognitive profiling suitable for assistive human robot~\cite{holzinger2019causability}.

\subsubsection{Severity Index}

A deterministic severity index was defined as:
Let $P_k(\text{AD})$ denote the predicted AD probability from fold $k$. We define $\overline{P(\text{AD})}$ as the mean predicted probability across $K$ folds:
\[
\overline{P(\text{AD})} = \frac{1}{K} \sum_{k=1}^{K} P_k(\text{AD}).
\]
\[
\text{SeverityIndex} = \alpha \cdot \overline{P(\text{AD})}
+ \beta \cdot \text{VoteRate}
- \gamma \cdot \text{Var}(P)
\]
This formulation ensures reproducibility and avoids LLM-driven variability in risk assessment~\cite{gibbons2012continuous}.

\subsection{Step 2: LLM-Based Assistive Feature Planning}

Only structured numerical summaries are passed to a local Qwen~2.5 (7B) LLM. Raw transcripts are never provided. The LLM is constrained to numeric inputs, produces structured JSON outputs, and provides non-diagnostic assistive recommendations. Supported features include Daily Reminder, Scheduler, Match the Fruit, XOX Game, and Memory Cues (photos/videos). Both the Patient LLM and Categorizer LLM used in this work are off the shelf large language models and were not fine tuned on medical, clinical, or Alzheimer's specific datasets. All task adaptation was achieved exclusively through prompt based in-context learning. This design choice was intentional to avoid embedding domain specific clinical priors into the system and to ensure that all stage conditioning behavior arises transparently from prompt structure rather than latent medical knowledge.

We defined three stage profiles aligned with mild-to-moderate impairment levels commonly referenced in HRI prototyping (Stage 1, Stage 3, Stage 5), and instantiated nine fictional personas (three per stage). A 10-item probe targeted episodic, prospective, working/short-term, semantic, and sequencing domains (two items each). A ``Patient LLM'' generated persona-conditioned answers, and a separate ``Categorizer LLM'' assigned one primary domain per item, flagged error severity, aggregated per-domain error totals, and produced a coarse stage estimate. Three personas were used as anchors for qualitative calibration and six evaluation.

\section{Results}

\subsection{Cross-Validation Performance}

Table~\ref{tab:cv_performance} reports five-fold cross-validation results for all tasks. The Cookie Theft task, which includes real Alzheimer's disease (AD) patients and real healthy controls, achieves balanced performance with an AUC of 0.878 and specificity exceeding sensitivity. This indicates that the model is more conservative in assigning AD labels, reducing false positives while maintaining reasonable detection sensitivity.

\begin{table}[h]
\centering
\small
\caption{Five-fold cross-validation performance. Perfect separation in tasks using synthetic healthy controls reflects distributional artifacts rather than real-world generalization.}
\setlength{\tabcolsep}{1pt}
\renewcommand{\arraystretch}{0.9}
\begin{tabular}{lccccc}
\hline
\textbf{Task} & \textbf{Acc.} & \textbf{F1} & \textbf{AUC} & \textbf{Sens.} & \textbf{Spec.} \\
\hline
Cookie & $0.792\!\pm\!0.038$ & $0.797\!\pm\!0.049$ & $0.878\!\pm\!0.045$ & $0.741\!\pm\!0.082$ & $0.856\!\pm\!0.057$ \\
Recall & 1.000 & 1.000 & 1.000 & 1.000 & 1.000 \\
Fluency & 1.000 & 1.000 & 1.000 & 1.000 & 1.000 \\
Sentence & 1.000 & 1.000 & 1.000 & 1.000 & 1.000 \\
\hline
\end{tabular}
\label{tab:cv_performance}
\end{table}

In contrast, perfect separation in Recall, Fluency, and Sentence tasks reflects the use of synthetically generated healthy controls, which likely introduce distributional artifacts. These results are therefore interpreted as feasibility validation rather than evidence of clinical generalization.

\subsection{Confusion Matrix Analysis}

\begin{figure}[h]
\centering
\includegraphics[width=\columnwidth]{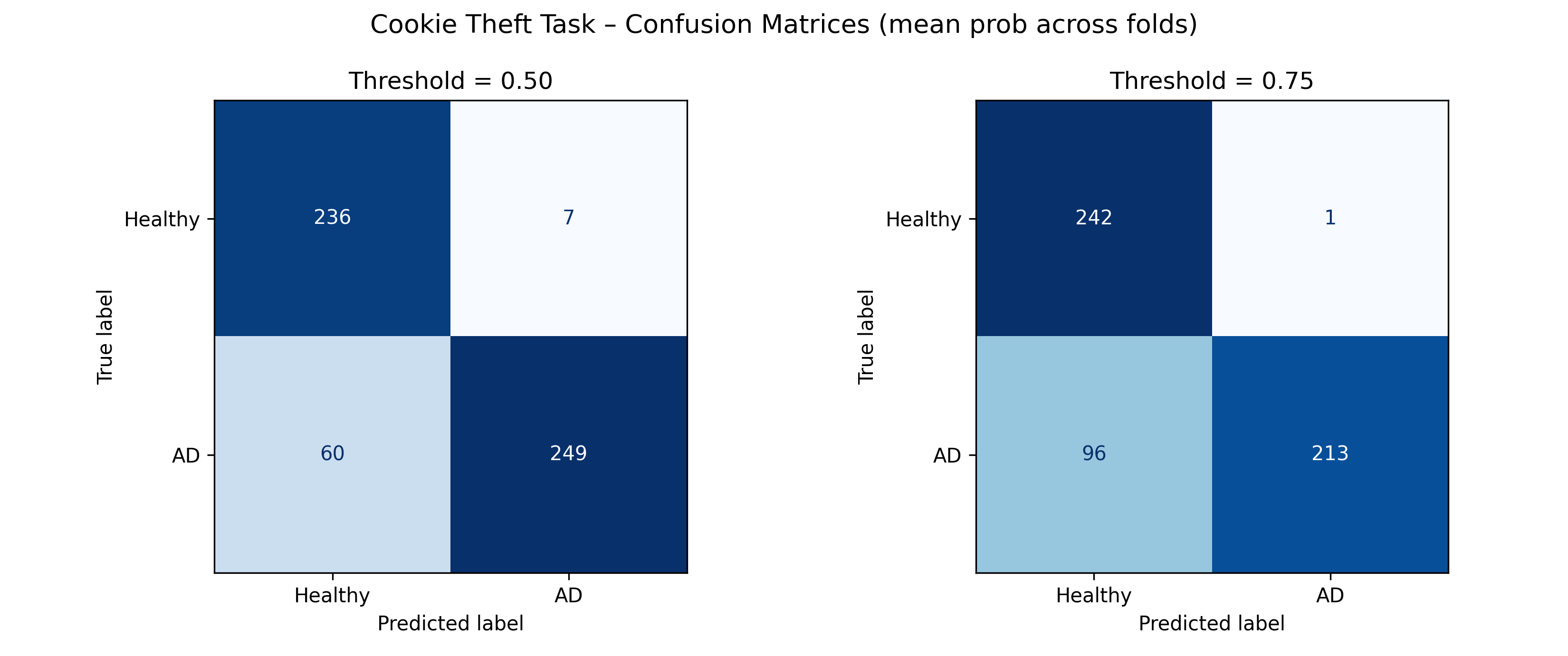}
\caption{Confusion matrices for Cookie Theft classification at different decision thresholds.}
\label{fig:confusion}
\end{figure}

Figure~\ref{fig:confusion} shows confusion matrices for the Cookie Theft task at thresholds 0.50 and 0.75. At a threshold of 0.50, the model demonstrates balanced classification behavior with moderate false negatives and low false positives. Increasing the threshold to 0.75 improves specificity while reducing sensitivity, illustrating a controllable trade-off between minimizing false alarms and avoiding missed detections. This behavior supports the use of threshold calibration in assistive, non-diagnostic settings where conservative predictions may be preferable.

\subsection{Severity Distribution}

\begin{figure}[h]
\centering
\includegraphics[width=0.9\columnwidth]{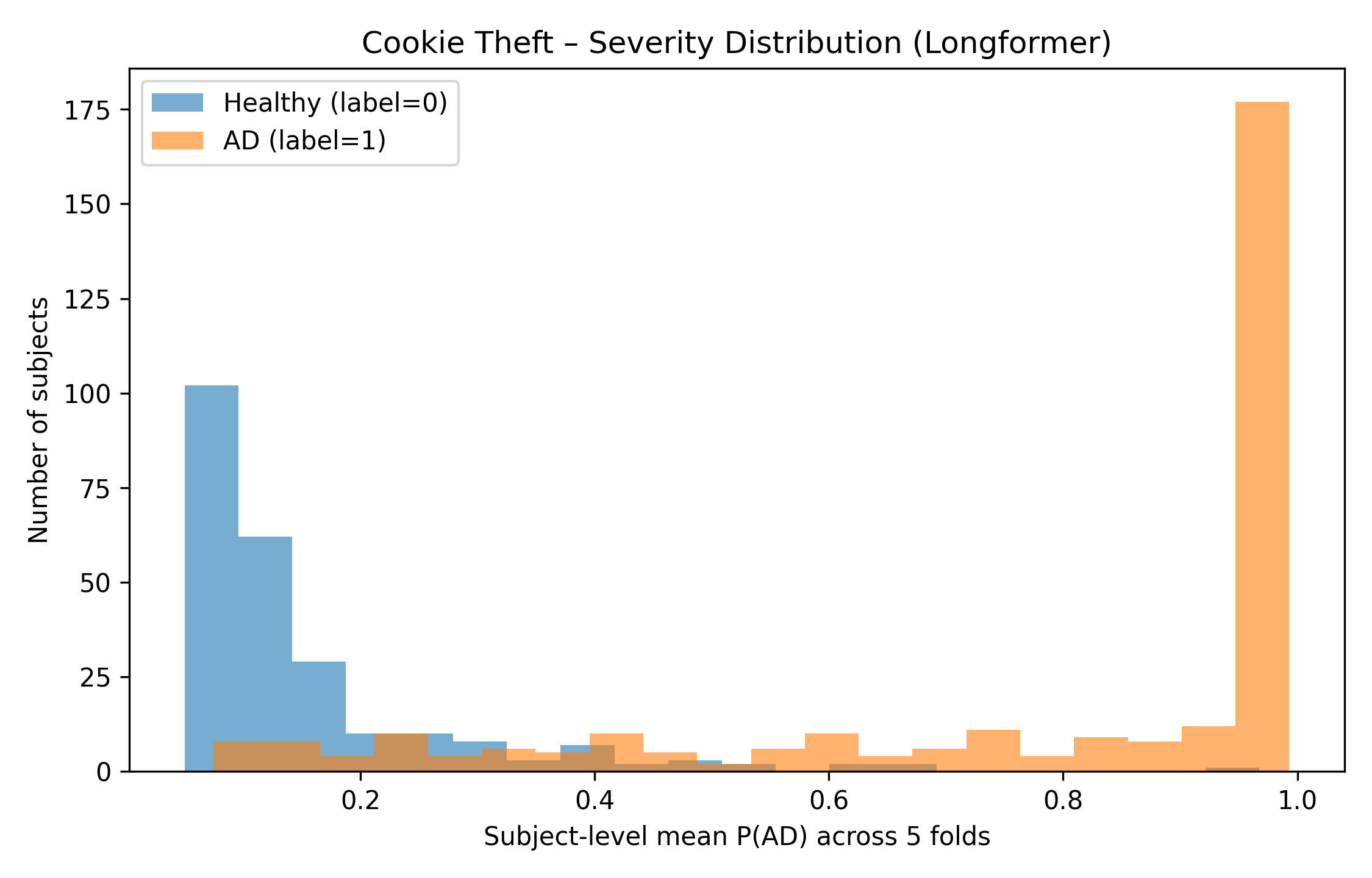}
\caption{Distribution of subject-level mean AD probabilities across five folds.}
\label{fig:severity}
\end{figure}

Figure~\ref{fig:severity} presents the distribution of subject-level mean AD probabilities aggregated across folds. Healthy subjects cluster predominantly at lower probability values, while AD subjects occupy higher probability regions with broader dispersion.

This separation indicates that the severity signal is not binary but continuous, enabling graded interpretation of cognitive risk rather than strict classification. Such continuous scoring supports stage-aware assistive adaptation in MEMOR-E.

\subsection{Cross-Fold Stability}

\begin{figure}[h]
\centering
\includegraphics[width=\columnwidth]{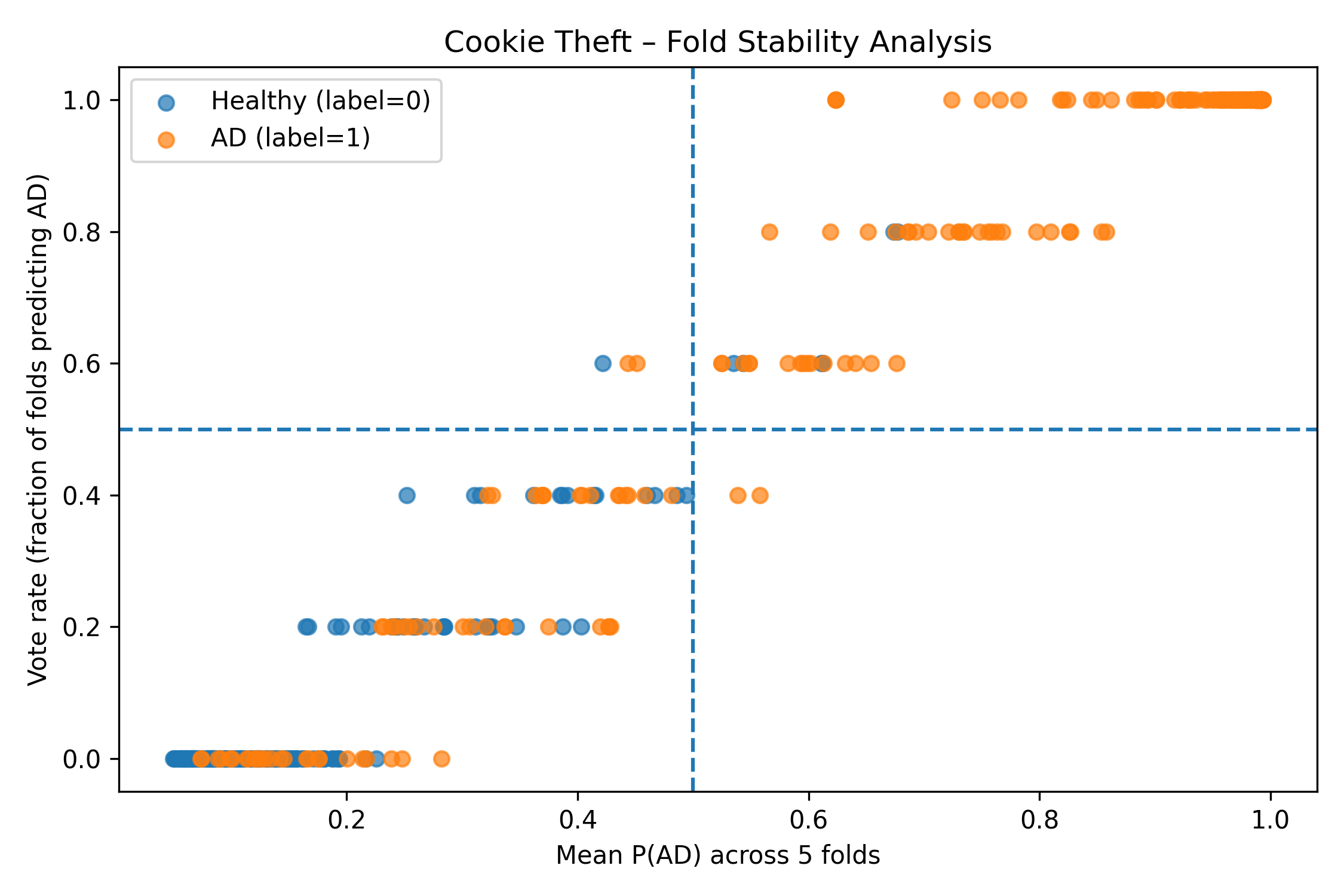}
\caption{Vote rate versus mean AD probability across folds.}
\label{fig:stability}
\end{figure}

Figure~\ref{fig:stability} illustrates prediction stability by plotting vote rate against mean AD probability across folds. AD subjects cluster in regions of both high mean probability and high vote rate, indicating stable cross-fold agreement. Healthy subjects predominantly occupy low-probability, low-vote-rate regions. Intermediate cases exhibit lower vote consistency, highlighting the incorporating cross-fold variance into the severity index to avoid unstable predictions.

\subsection{Clinically Grounded Attribution Analysis}

To examine whether the model relies on linguistically meaningful discourse markers, we performed a BPE-aware rebucketing of token attributions and compared both normalized attribution mass and raw token frequency across clinically relevant categories.

\begin{figure}[h]
\centering
\includegraphics[width=\columnwidth]{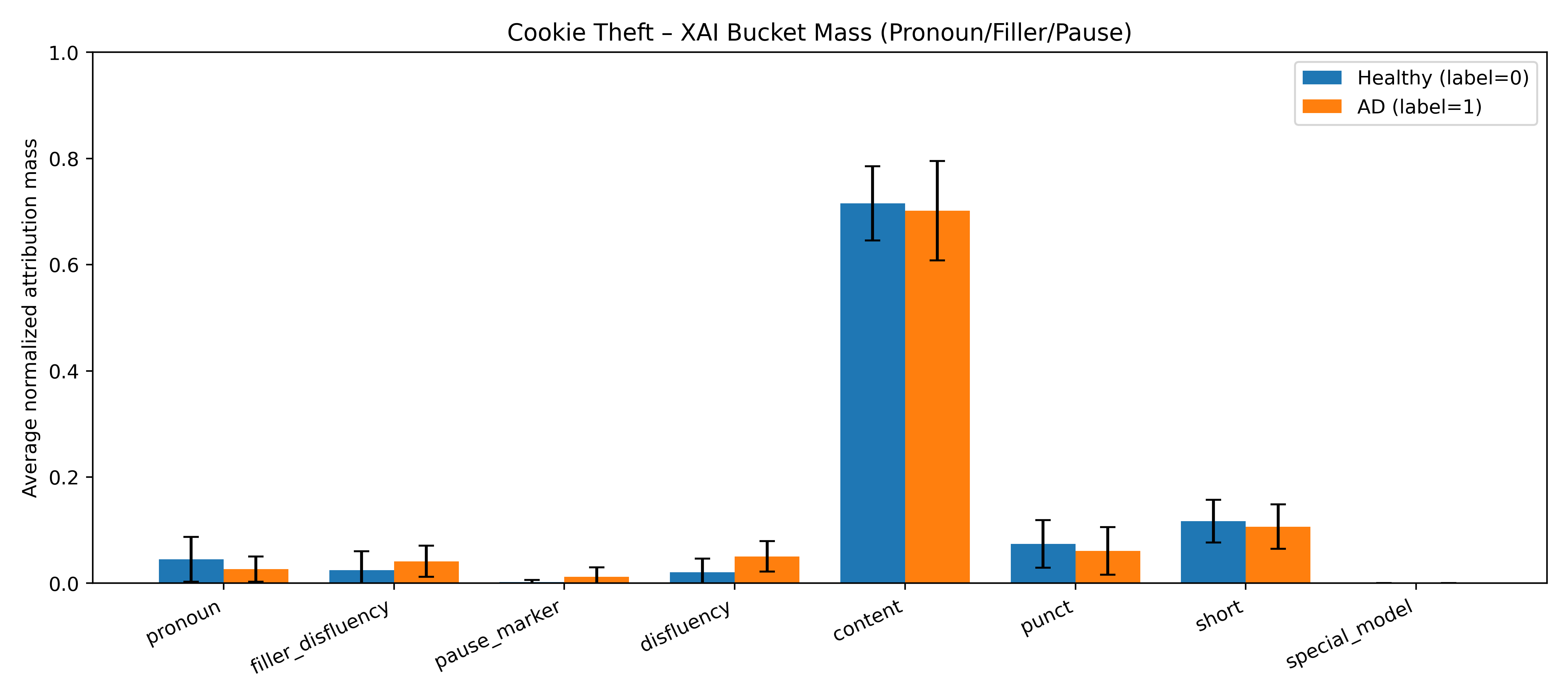}
\caption{Normalized attribution bucket mass after BPE-aware rebucketing across all folds. AD predictions exhibit elevated attribution mass on disfluency-related categories.}
\label{fig:xai_bucket}
\end{figure}

As shown in Figure~\ref{fig:xai_bucket}, lexical content tokens constitute the dominant attribution source for both groups (approximately 70\% of normalized mass), indicating that semantic information remains the primary discriminative signal. This suggests that the classifier does not rely on superficial artifacts but instead bases its decisions largely on meaningful lexical content.

Beyond this dominant semantic signal, systematic differences emerge in clinically relevant discourse markers. Filler disfluencies (e.g., ``uh'', ``um'') and pause markers (e.g., ``(.)'', ``(..)'') were more frequent in AD speech and received substantially higher attribution mass during AD classification. Specifically, filler tokens accounted for 6.38\% of AD tokens compared to 5.04\% in Healthy speech, with attribution mass elevated in AD predictions (4.15\% vs 2.49\%). Similarly, pause markers were more prevalent in AD (1.48\% vs 0.72\%) and received higher attribution mass (1.24\% vs 0.24\%).

In contrast, pronoun tokens were more frequent in Healthy controls (4.18\% vs 3.43\%) and received higher attribution mass during Healthy predictions (4.51\% vs 2.66\%). This indicates that the classifier associates pronoun-rich, structurally coherent discourse patterns with non-pathological speech.

Importantly, the directional alignment between token frequency differences and attribution mass differences suggests that the model internalizes genuine linguistic distributional patterns rather than exploiting spurious correlations. Model special tokens contributed negligible attribution mass, further confirming that classification decisions are not driven by tokenization artifacts.

Taken together, these findings indicate that the classifier primarily relies on semantic content while incorporating disfluency related signals as secondary discriminative cues. The observed attribution patterns are consistent with established descriptions of hesitation and fluency alterations in Alzheimer's discourse, supporting the interpretability and cognitive plausibility of the model's internal decision processes.
To interpret the dominance of lexical content attributions, we examined high attribution open class words driving classification decisions. AD predictions primarily emphasized concrete nouns and action descriptors related to picture content, often with reduced specificity or incomplete referential grounding. In contrast, Healthy predictions showed higher attribution mass on structurally cohesive lexical sequences with consistent noun--verb relations and temporal organization. This indicates that discriminative behavior arises from the organization and contextual integration of lexical content rather than its mere presence, aligning with established discourse level degradation patterns in Alzheimer's speech and supporting the cognitive plausibility of the model's attributions.

\subsection{Complementary Feasibility Results on Alzheimer's Stage-Conditioned Personas}
\label{subsec:persona_results}

To complement dataset-driven evaluation, we report results from an earlier feasibility study examining whether a stage-conditioned \emph{Patient LLM} produces responses consistent with increasing cognitive impairment, and whether a separate \emph{Categorizer LLM} converts question--answer interactions into interpretable domain summaries and a coarse stage estimate. These results serve as a prototype validation of interpretability and stage awareness.

The Patient LLM exhibits a monotonic increase in total errors from Stage~1 to Stage~5, suggesting that conditioning prompts can induce stage-consistent degradation trends. Domain-level patterns also qualitatively align with expectations, with more frequent working-memory, episodic/prospective, and sequencing errors under higher impairment prompts while retaining semantic facts.

\subsubsection{Categorizer LLM: Domain Summaries and Stage Estimation}
We evaluated the Categorizer LLM by comparing predicted profiles against ground-truth stage anchors, reporting per-domain absolute errors and total categorization error across six evaluation personas.

\section{Discussion}

The results indicate that MEMOR-E can support interpretable, Alzheimer's stage-aware cognitive summaries that are valuable for adaptive interaction and caregiver awareness. Importantly, the system is not intended to replace clinical assessment but to augment daily support through context aware, non diagnostic~\cite{yang2020caregiver}. The robot's mobility enables point of need interaction, which static assistive devices cannot provide, while explainable language based reasoning supports transparency and trust in human-robot interaction~\cite{topol2019highperformance}.
In addition to DementiaBank-based transcript modeling, the persona-based feasibility results suggest that LLM-driven stage conditioning and error summarization can provide interpretable, domain-level signals that complement dataset-driven classifiers when designing assistive (non-diagnostic) interaction policies. MEMOR-E is designed for safety-critical environments involving cognitively vulnerable users. The current system operates under conservative autonomy assumptions: all assistive actions are non-invasive, reversible, and mediated through a tablet interface. Navigation relies on established ROS 2 safety stacks, and no physical manipulation or medication dispensing is performed. As future controlled studies are conducted, trustworthiness metrics including caregiver override frequency, false alarm rates, interaction predictability, and user distress indicators will be incorporated. These measures will complement existing explainability guarantees to support safe, transparent, and accountable deployment.

\section{Limitations and Future Work}

This work presents a feasibility oriented, privacy preserving, and explainable framework for cognitive profiling in assistive robotics. Limitations include reliance on synthetic controls for most tasks, persona-driven simulations, and pre collected transcripts, which constrain ecological validity and real-world generalization. Attribution based explanations and severity indices are model dependent and not clinically validated, requiring additional safeguards and human oversight. Future work will focus on collecting real control data, conducting longitudinal and live studies, integrating multimodal sensing, and strengthening clinical validation and human in the loop deployment. While the current study focuses on feasibility and interpretability, caregiver perspectives are critical for real-world adoption. Future work will include qualitative and mixed-method user studies with caregivers to evaluate perceived usefulness, cognitive workload reduction, trust in explainable summaries, and impact on daily care routines. These studies will assess how MEMOR-E augments not replaces human caregiving, aligning system behavior with practical care needs.

\section{Conclusion}

This paper presented MEMOR-E, a privacy preserving and explainable framework for cognitively adaptive assistive robotics in Alzheimer's care. The system combines task specific Longformer classifiers, attribution-based explainable AI (XAI)~\cite{zhang2021edge}, and a locally deployed large language model on a mobile quadruped platform. Language derived cognitive signals are transformed into structured, interpretable, non diagnostic assistive insights. Cookie Theft results using real Alzheimer's disease patients and healthy controls show stable severity trends and cognitively plausible attributions. Attribution bucket statistics and a deterministic severity index provide transparent characterization of cognitive difficulty while preserving transcript privacy~\cite{leite2013longterm}.

MEMOR-E employs a pipeline that relies exclusively on XAI-derived statistical summaries for assistive planning. The framework is stage-aware and non-diagnostic, integrating socially assistive robotics, long-context language modeling, and explainable AI within ethical and human-centered constraints. Future work will focus on real-world deployment and clinical.

\section*{Acknowledgements}
This study uses data from the DementiaBank Pitt corpus provided by the TalkBank project~\cite{macwhinney2011talkbank, becker1994natural}.
We acknowledge support from the National Institute on Aging (NIA) grants AG03705 and AG05133.

\newpage
\appendix
\section*{Appendix}

\section{In-Context Learning for Stage Conditioned Personas and LLM Categorization}
\label{app:icl}

To complement the dataset-driven evaluation, we include results from an earlier feasibility study that assessed whether LLMs can (i) emulate stage-consistent cognitive response patterns and (ii) convert raw question--answer interactions into interpretable domain- and severity-level summaries suitable for non-diagnostic assistive planning.
Example demonstrations are shown below.

\begin{figure}[h]
\begin{tcolorbox}[colback=gray!10, colframe=gray!60, boxrule=0.4pt]

\textbf{Example 1 (demonstration)}

\textbf{Input features:}

severity\_index = 0.72 \\
vote\_rate = 0.80 \\
variance = 0.04 \\
disfluency\_ratio = 0.31 \\
content\_mass = 0.68

\textbf{Model output:}

Assistive features: Daily Reminder, Memory Cues, Match-the-Fruit cognitive game

\medskip

\textbf{Example 2 (demonstration)}

\textbf{Input features:}

severity\_index = 0.38 \\
vote\_rate = 0.40 \\
variance = 0.02 \\
disfluency\_ratio = 0.12 \\
content\_mass = 0.81

\textbf{Model output:}

Assistive features: Routine reminders, light cognitive games, optional scheduling support.

\medskip

\textbf{New Input (query)}

severity\_index = 0.55, vote\_rate = 0.63, variance = 0.03 \\
disfluency\_ratio = 0.21, content\_mass = 0.74

\textbf{Task:} Recommend appropriate assistive interaction features.

\end{tcolorbox}
\caption{Example of prompt-based in-context learning used to map explainable cognitive statistics to assistive feature recommendations. Demonstration examples condition the LLM before generating recommendations for a new input.}
\label{fig:icl_example}
\end{figure}

These demonstrations condition the LLM to produce structured assistive recommendations based solely on explainable statistical summaries without accessing raw patient transcripts.

\section{Detailed Attribution Analysis}
\label{app:attrib}

To examine the linguistic signals used by the Longformer classifier, we analyzed token-level attributions produced by Integrated Gradients. Subword tokens produced by byte-pair encoding (BPE) were reconstructed into word-level units prior to analysis.

Tokens were grouped into linguistically meaningful categories to enable privacy-preserving aggregation:

\begin{itemize}
\item Pronouns
\item Fillers and disfluency markers
\item Pause annotations
\item Lexical content tokens
\item Punctuation and special tokens
\end{itemize}

Table~\ref{tab:attrib} reports both token frequency and normalized attribution mass across discourse categories.

\begin{table}[h]
\centering
\caption{Token frequency and normalized attribution mass by discourse category (Cookie Theft task).}
\label{tab:attrib}
\begin{tabular}{lcccc}
\hline
Feature & Freq(HC) & Freq(AD) & Attr(HC) & Attr(AD) \\
\hline
Pronoun & 4.18\% & 3.43\% & 4.51\% & 2.66\% \\
Filler & 5.04\% & 6.38\% & 2.49\% & 4.15\% \\
Pause & 0.72\% & 1.48\% & 0.24\% & 1.24\% \\
\hline
\end{tabular}
\end{table}

The analysis indicates that lexical content tokens dominate attribution mass, representing approximately 70\% of the model's evidence during prediction. However, disfluency markers such as fillers and pauses contribute proportionally more to Alzheimer's disease (AD) predictions than to healthy control predictions. This pattern aligns with known characteristics of Alzheimer's speech, including increased hesitation and reduced fluency.

\section{Persona-Based Feasibility Study Results}
\label{app:persona}

To complement dataset-based evaluation, we conducted a feasibility study using stage-conditioned fictional personas generated by a large language model. Personas were designed to emulate cognitive behavior associated with three levels of Alzheimer's impairment: Stage 1 (mild), Stage 3 (moderate), and Stage 5 (moderately severe).

\begin{figure}[H]
  \centering
  \includegraphics[width=1\columnwidth]{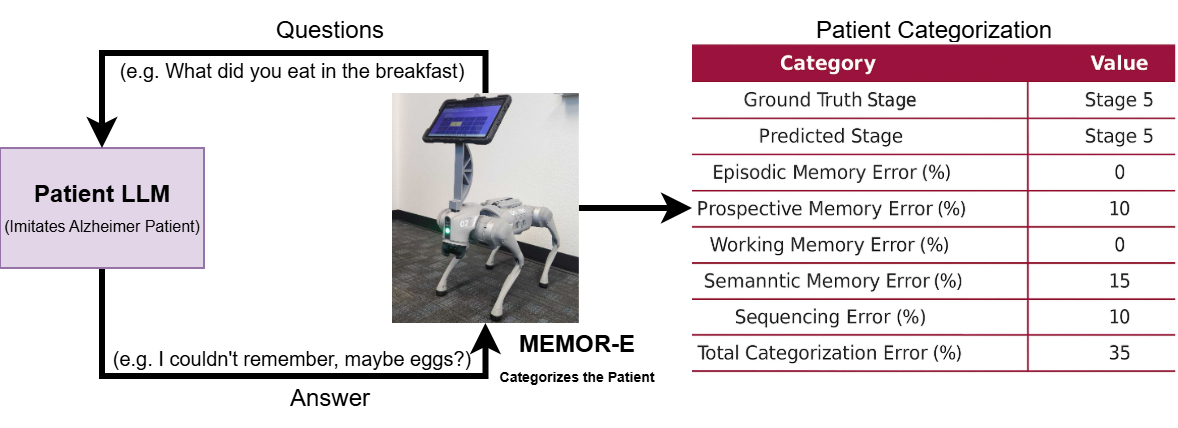}
  \caption{Patient LLM and the categorizer LLM.}
  \label{fig:patient_categorizer_LLM}
\end{figure}

Each persona responded to a set of ten questions covering five cognitive domains:

\begin{itemize}
\item Episodic memory
\item Prospective memory
\item Working/short-term memory
\item Semantic knowledge
\item Sequential reasoning
\end{itemize}

Responses were scored as:
\begin{itemize}
\item 0 = correct
\item 0.5 = correct but uncertain
\item 1 = incorrect
\end{itemize}

Table~\ref{tab:stage_errors} summarizes the average percentage of incorrect responses across stages.

\begin{table}[H]
\centering
\caption{Average error percentages across stage-conditioned personas.}
\label{tab:stage_errors}
\begin{tabular}{lccccc}
\hline
Stage & Epis. & Pros. & Work. & Sem. & Seq. \\
\hline
Stage 1 & 5.00\% & 5.00\% & 6.67\% & 6.67\% & 10.00\% \\
Stage 3 & 13.33\% & 13.33\% & 16.67\% & 3.33\% & 10.00\% \\
Stage 5 & 13.33\% & 15.00\% & 16.67\% & 8.33\% & 11.67\% \\
\hline
\end{tabular}
\end{table}

Results show a monotonic increase in total errors from Stage 1 to Stage 5. We further evaluated a separate Categorizer LLM responsible for converting question--answer interactions into domain summaries and stage estimates.

\begin{table}[H]
\centering
\small
\setlength{\tabcolsep}{3pt}
\caption{Categorizer LLM evaluation across six personas.}
\label{tab:categorizer}
\begin{tabular}{lllcccr}
\hline
Patient & GT Stage & Pred Stage & $\Delta$Epis & $\Delta$Pros & $\Delta$Work & Cat.Err \\
\hline
P1 & Stage 1 & Stage 3 & 0\% & 0\% & 0\% & 5\% \\
P2 & Stage 1 & Stage 3 & 5\% & 0\% & 5\% & 10\% \\
P3 & Stage 3 & Stage 3 & 5\% & 0\% & 0\% & 10\% \\
P4 & Stage 3 & Stage 3 & 0\% & 5\% & 5\% & 15\% \\
P5 & Stage 5 & Stage 5 & 0\% & 10\% & 0\% & 35\% \\
P6 & Stage 5 & Stage 5 & 0\% & 10\% & 10\% & 25\% \\
\hline
\end{tabular}
\end{table}

Across the six evaluation personas, the Categorizer LLM achieved a stage prediction accuracy of 66.7\%. While these results reflect a simulated setting, they demonstrate that prompt-conditioned language models can produce interpretable cognitive summaries aligned with stage-level behavioral patterns.

\balance

\begin{thebibliography}{10}

\bibitem{jack2018framework}
Clifford~R. Jack et~al.
\newblock Nia-aa research framework: Toward a biological definition of alzheimer’s disease.
\newblock {\em Alzheimer’s \& Dementia}, 14(4):535--562, 2018.

\bibitem{bemelmans2012scoping}
R.~Bemelmans, G.~J. Gelderblom, P.~Jonker, and L.~de~Witte.
\newblock Socially assistive robots in elderly care: A systematic review.
\newblock {\em Journal of Medical Internet Research}, 14(3), 2012.

\bibitem{rossi2017users}
Silvia Rossi, Mariacarla Staffa, and Guglielmo Tamburrini.
\newblock Users’ trust in socially assistive robots.
\newblock In {\em Proceedings of the ACM/IEEE International Conference on Human-Robot Interaction (HRI)}, 2017.

\bibitem{chapman1995narrative}
Sandra~B. Chapman, Hanna~K. Ulatowska, and Laurie~R. Franklin.
\newblock Narrative discourse in alzheimer’s disease.
\newblock {\em Journal of Speech and Hearing Research}, 38:402--414, 1995.

\bibitem{perret1974fluency}
Edith Perret.
\newblock The left frontal lobe and the suppression of habitual responses.
\newblock {\em Neuropsychologia}, 12:323--330, 1974.

\bibitem{becker1994natural}
James~T. Becker, Francois Boller, Oscar~L. Lopez, Julie Saxton, and Karen~L. McGonigle.
\newblock The natural history of alzheimer's disease: Description of study cohort and accuracy of diagnosis.
\newblock {\em Archives of Neurology}, 51(6):585--594, 1994.

\bibitem{lehman2021bias}
Eric Lehman et~al.
\newblock Does bert pretraining on clinical notes encode health disparities?
\newblock In {\em Proceedings of the 59th Annual Meeting of the Association for Computational Linguistics (ACL)}, 2021.

\bibitem{sundararajan2017axiomatic}
Mukund Sundararajan, Ankur Taly, and Qiqi Yan.
\newblock Axiomatic attribution for deep networks.
\newblock In {\em International conference on machine learning}, pages 3319--3328. PMLR, 2017.

\bibitem{holzinger2019causability}
Andreas Holzinger et~al.
\newblock Causability and explainability of artificial intelligence in medicine.
\newblock {\em ACM Computing Surveys}, 52(5), 2019.

\bibitem{gibbons2012continuous}
Laura~E. Gibbons et~al.
\newblock A composite score for executive functioning.
\newblock {\em Alzheimer Disease \& Associated Disorders}, 26(4):336--343, 2012.

\bibitem{yang2020caregiver}
Qian Yang et~al.
\newblock Investigating how clinicians use ai-assisted diagnosis tools.
\newblock In {\em Proceedings of the 2020 CHI Conference on Human Factors in Computing Systems}, 2020.

\bibitem{topol2019highperformance}
Eric Topol.
\newblock High-performance medicine: The convergence of human and artificial intelligence.
\newblock {\em Nature Medicine}, 25:44--56, 2019.

\bibitem{zhang2021edge}
Chi Zhang et~al.
\newblock Edge intelligence for healthcare.
\newblock {\em ACM Computing Surveys}, 2021.

\bibitem{leite2013longterm}
Iolanda Leite, Carlos Martinho, and Ana Paiva.
\newblock Long-term human–robot interaction.
\newblock {\em International Journal of Social Robotics}, 5:291--308, 2013.

\bibitem{macwhinney2011talkbank}
Brian MacWhinney.
\newblock The {TalkBank} project.
\newblock {\em Journal of Psycholinguistic Research}, 40(2):149--156, 2011.

\end{thebibliography}
\end{document}